\documentclass[lettersize,journal]{IEEEtran}
\usepackage{amsmath,amsfonts}
\usepackage{algorithm}
\usepackage{algpseudocode}
\usepackage{array}
\usepackage[caption=false,font=normalsize,labelfont=sf,textfont=sf]{subfig}
\usepackage{textcomp}
\usepackage{stfloats}
\usepackage{url}
\usepackage{verbatim}
\usepackage{graphicx}
\usepackage{cite}
\usepackage{hyperref}
\usepackage{xcolor}
\usepackage{soul}
\sethlcolor{yellow} 
\usepackage{mathrsfs}
\usepackage{amssymb}
\begin{document}

\title{TriAdaptLoRA: Brain-Inspired Triangular Adaptive Low-Rank Adaptation for Parameter-Efficient Fine-Tuning}

\author{Yao Liang, Yuwei Wang, Yi Zeng

\thanks{Yao Liang is with the Brain-inspired Cognitive Intelligence Lab, Institute of Automation, Chinese Academy of Sciences, Beijing 100190, China, and School of Artificial Intelligence, University of Chinese Academy of Sciences, Beijing 100049, China.} 
\thanks{Yuwei Wang is with the Brain-inspired Cognitive Intelligence Lab, Institute of Automation, Chinese Academy of Sciences, Beijing 100190, China, and Center for Long-term Artificial Intelligence, Beijing 100190, China.}

\thanks{Yi Zeng is with the Brain-inspired Cognitive Intelligence Lab, Institute of Automation, Chinese Academy of Sciences, Beijing 100190, China, and Center for Long-term Artificial Intelligence, Beijing 100190, China, and University of Chinese Academy of Sciences, Beijing 100083, China, and Key Laboratory of Brain Cognition and Brain-inspired Intelligence Technology, Chinese Academy of Sciences, Shanghai, 200031, China.}

\thanks{The first and the second authors contributed equally to this work, and serve as co-first authors.}
\thanks{The corresponding author is Yi Zeng (e-mail: yi.zeng@ia.ac.cn).}}

\maketitle
\begin{abstract}
The fine-tuning of Large Language Models (LLMs) is pivotal for achieving optimal performance across diverse downstream tasks. However, while full fine-tuning delivers superior results, it entails significant computational and resource costs. Parameter-Efficient Fine-Tuning (PEFT) methods, such as LoRA, address these challenges by reducing the number of trainable parameters, but they often struggle with rank adjustment efficiency and task-specific adaptability.
We propose Triangular Adaptive Low-Rank Adaptation (TriAdaptLoRA), a novel PEFT framework inspired by neuroscience principles, which dynamically optimizes the allocation of trainable parameters. TriAdaptLoRA introduces three key innovations: 1) a triangular split of transformation matrices into lower and upper triangular components to maximize parameter utilization, 2) a parameter importance metric based on normalized Frobenius norms for efficient adaptation, and 3) an adaptive rank-growth strategy governed by dynamic thresholds, allowing flexible parameter allocation across training steps.
Experiments conducted on a variety of natural language understanding and generation tasks demonstrate that TriAdaptLoRA consistently outperforms existing PEFT methods. It achieves superior performance, enhanced stability, and reduced computational overhead, particularly under linear threshold-driven rank growth. These results highlight its efficacy as a scalable and resource-efficient solution for fine-tuning LLMs.
\end{abstract}


\section{Introduction}

In recent years, large language models (LLMs) have emerged as foundational tools across a wide spectrum of downstream tasks, delivering exceptional performance and driving both academic and industrial innovation~\cite{ouyang2022training,ding2023parameter,ziems-etal-2023-large}. Flagship models such as GPT-4~\cite{achiam2023gpt} dominate closed-source development, while Llama3~\cite{dubey2024llama} leads progress in the open-source ecosystem. However, training these models from scratch requires vast computational resources. For example, training Llama2~\cite{touvron2023llama2} demands approximately 3.3 million GPU hours on an NVIDIA A100-80GB graphics card, resulting in an estimated 539 tons of carbon emissions. Such resource-intensive processes underscore the need for more sustainable solutions, especially given the performance gains associated with scaling up LLMs as observed in OpenAI’s power-law scaling studies~\cite{askell2021general}.

Fine-tuning has become a critical strategy for adapting pre-trained LLMs to specific tasks~\cite{wei2021finetuned,min2021metaicl,liu2022few,ouyang2022training}. While full fine-tuning provides high performance by updating all model parameters, it remains prohibitively expensive in terms of computational and memory costs, limiting its practical applicability~\cite{qiu2020pre,raffel2020exploring}. 

To address these challenges, PEFT methods have gained significant attention, with Low-Rank Adaptation (LoRA)~\cite{hu2022lora} emerging as one of the most widely adopted approaches. LoRA achieves efficiency by freezing the pre-trained model's parameters and introducing low-rank matrices for adaptation. However, LoRA's fixed-rank configuration for all low-rank matrices limits its adaptability and parameter utilization efficiency. 

Efforts to overcome these limitations have driven the development of three main categories of LoRA-based improvements:
\begin{itemize}
    \item \textbf{Structural optimization methods}: These include approaches like DoRA~\cite{liu2024dora}, VeRA~\cite{kopiczko_vera_2024}, AFLoRA~\cite{liu_aflora_2024}, and PRoLoRA~\cite{wang_prolora_2024}, which enhance LoRA’s structure through mechanisms such as matrix decomposition, parameter sharing, and gradual freezing, improving parameter efficiency and performance.
    \item \textbf{Application-oriented methods}: Examples include LongLoRA~\cite{chen2023longlora}, QLoRA~\cite{dettmers2024qlora}, and LoRA-Flow~\cite{wang_lora-flow_2024}, which extend LoRA's applicability to contexts such as long-document processing, quantization, and multi-module integration.
    \item \textbf{Rank-adjustment optimization methods}: These dynamically adjust the ranks of low-rank matrices for greater efficiency and task-specific performance. Prominent examples include AdaLoRA~\cite{zhang2023adaptive}, IncreLoRA~\cite{Zhang2023IncreLoRAIP}, and CAPABOOST~\cite{song2024increasing}. While these methods address certain limitations of LoRA, they introduce challenges such as high memory requirements during initialization, significant computational overhead, and inefficiencies stemming from fixed thresholds in rank adjustments.
\end{itemize}

Motivated by these challenges, we introduce TriAdaptLoRA, a novel PEFT method inspired by synaptic plasticity mechanisms in the brain, particularly Hebbian learning~\cite{hebb_organization_2005}. TriAdaptLoRA dynamically allocates trainable parameters to incremental matrices based on their importance, optimizing rank distribution and parameter efficiency. By combining triangular split and adaptive rank growth strategies, our approach not only enhances parameter utilization but also significantly reduces computational overhead compared to existing methods.

To validate TriAdaptLoRA, we conducted extensive experiments on natural language understanding and generation tasks, demonstrating its superior performance and efficiency compared to existing methods. Specifically, our method achieves notable improvements on the GLUE benchmark~\cite{wang2018glue} and SQuAD 2.0~\cite{rajpurkar-etal-2018-know} tasks while maintaining low computational costs.

The main contributions of this paper are as follows:
\begin{itemize}
    \item \textbf{Efficient parameter expansion}: We propose a novel triangular split structure that enhances the scalability and performance of parameter-efficient fine-tuning.
    \item \textbf{Importance-driven parameter allocation}: We develop an efficient algorithm to evaluate the importance of incremental matrices, reducing the computational complexity of rank adjustment compared to AdaLoRA and IncreLoRA.
    \item \textbf{Dynamic rank growth}: We introduce an adaptive mechanism for dynamically determining the number of incremental matrices participating in rank expansion, mitigating inefficiencies associated with fixed thresholds in traditional methods.
    \item \textbf{Hyperparameter reduction}: Unlike AdaLoRA and IncreLoRA, TriAdaptLoRA avoids the need for additional hyperparameters such as sensitivity smoothing and final warm-up steps, simplifying the fine-tuning process and reducing computational costs.
\end{itemize}

These contributions position TriAdaptLoRA as a robust and efficient solution for fine-tuning LLMs in resource-constrained settings.

\section{Related work}

LLMs such as GPT-4~\cite{achiam2023gpt} and Llama3~\cite{dubey2024llama} have achieved significant progress in natural language processing and multimodal tasks. These models are typically based on the Transformer architecture~\cite{vaswani2017attention}, whose core component is the multi-head self-attention mechanism. This mechanism effectively captures dependencies between different positions in a sequence. The specific algorithm is illustrated in Equation \ref{eq1}:
\begin{equation}\label{eq1}
\begin{aligned}
\text{MultiHead}(Q, K, V) &= \text{Concat}(head_1, \ldots, head_h)W^O, \\
head_i &= \text{Attention}(QW_i^Q, KW_i^K, VW_i^V), \\
\text{Attention}(Q, K, V) &= \text{softmax}\left(\frac{QK^T}{\sqrt{d_k}}\right)V,
\end{aligned}
\end{equation}
where \(Q\), \(K\), and \(V\) represent Query, Key, and Value vectors, with \(W_i^Q\), \(W_i^K\), \(W_i^V\) as projection matrices and \(W^O\) as the output projection matrix, while \(d_k\) scales the dot product~\cite{vaswani2017attention}.

LoRA is one of the most representative PEFT methods to date~\cite{hu2022lora}. LoRA approximates the incremental matrix \(\Delta W\) of the pre-trained weight matrix \(W_0\) as the product of low-rank matrices \(A\) and \(B\). During training, the weight matrix \(W_0\) is frozen, and only the low-rank matrices \(A\) and \(B\) are updated, significantly reducing the number of trainable parameters while preserving model performance. Its forward propagation expression is Equation \ref{eq2}:
\begin{equation}\label{eq2}
\begin{aligned}
h = W_0 x + \Delta W x = W_0 x + B A x,
\end{aligned}
\end{equation}
where \( W_0 \in \mathbb{R}^{d \times n} \), \( A \in \mathbb{R}^{r \times n} \), and \( B \in \mathbb{R}^{d \times r} \). 
Compared to full fine-tuning, LoRA reduces the number of training parameters by several orders of magnitude when fine-tuning models like GPT-3 175B. However, LoRA still exhibits a performance gap compared to full fine-tuning. 
Additionally, the fixed and consistent rank of the low-rank matrices in LoRA limits the parameter efficiency and adaptability of the model. 
Building on LoRA, numerous improved methods have been developed to further enhance parameter efficiency and performance, particularly optimization methods for rank adjustment, like:

\begin{itemize}
\item AdaLoRA~\cite{zhang2023adaptive} employs singular value decomposition to approximate the incremental matrix and dynamically prunes based on the importance scores of vector groups within the incremental matrix. This approach dynamically adjusts the rank of the incremental matrices at different positions, thereby enhancing parameter efficiency and performance. 
\item IncreLoRA~\cite{Zhang2023IncreLoRAIP} introduces a rank self-growth mechanism for the low-rank matrices, eliminating the upper limit on the rank of the incremental matrices and avoiding additional memory requirements during initialization. 
\item Dynamic LoRA~\cite{valipour_dylora_2023} employs a random search approach for ranks, training multiple ranks for the incremental matrices simultaneously during the training process to facilitate rapid search for the optimal rank during testing. 
\item CAPABOOST~\cite{song2024increasing} enhances the rank of the incremental matrices by linearly combining multiple parallel low-rank matrices, thereby increasing the model's expressive power without adding extra parameters. 
\end{itemize}

\section{The Proposed Method}
\begin{figure*}[htb]
  \centering
  \includegraphics [width=17cm]{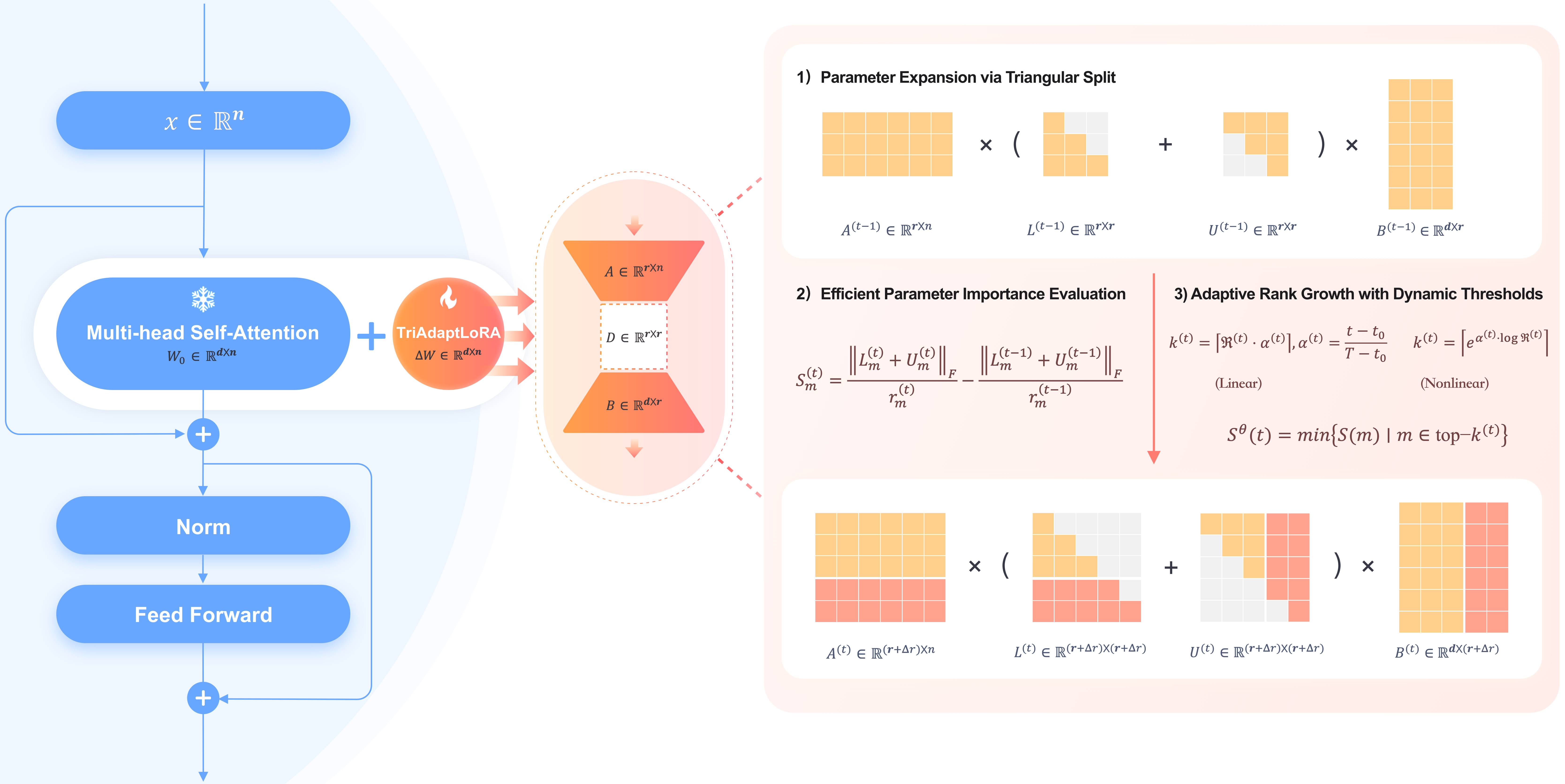}
  \caption{Schematic diagram of TriAdaptLoRA's overall structure.}
  \label{P1}
\end{figure*}

In this section, we present TriAdaptLoRA, a novel parameter-efficient fine-tuning method inspired by neuroscience principles and designed to optimize the allocation of trainable parameters dynamically. The approach enhances parameter utilization by introducing three key innovations: (1) decomposing transformation matrices into the sum of lower and upper triangular matrices; (2) a novel parameter importance metric for incremental matrices; and (3) a dynamic rank-growth strategy guided by adaptive thresholds. The overall structure of TriAdaptLoRA is illustrated in Figure \ref{P1}. The overall information processing workflow of the TriAdaptLoRA is described in Algorithm \ref{alg:trilora}.

\subsection{Neuroscience Inspiration}

In neuroscience, the optimization of synaptic strengths is a critical mechanism for enhancing neural network performance. This process, often referred to as the credit assignment problem, focuses on identifying connections that significantly influence behavior~\cite{littman_reinforcement_2015,roelfsema_control_2018,sutton_reinforcement_2018}. Hebb's rule~\cite{hebb_organization_2005}, a foundational theory in neuroscience, states that the strength of a synaptic connection increases when the presynaptic and postsynaptic neurons are simultaneously active. Mathematically, this is expressed as:
\begin{equation}\label{eq3}
\Delta w_{i,j} = \beta \cdot f_i(a_i) \cdot f_j(a_j),
\end{equation}
where $\Delta w_{i,j}$ is the change in connection strength between neurons $i$ and $j$, $\beta$ is the learning rate, and $f_i(a_i)$ and $f_j(a_j)$ represent the activities of the presynaptic and postsynaptic neurons, respectively. This principle inspires the dynamic adjustment of trainable parameters in TriAdaptLoRA, where the importance of model parameters evolves adaptively based on their contribution to learning.

\subsection{Parameter Expansion via Triangular Split}

We approximate the incremental matrix $\Delta W$ of the pre-trained weight matrix $W_0$ as follows:
\begin{equation}\label{eq4}
\Delta W = BDA, \ \ D = L + U,
\end{equation}
where $B \in \mathbb{R}^{d \times r}$ and $A \in \mathbb{R}^{r \times n}$ are low-rank matrices, and $L, U \in \mathbb{R}^{r \times r}$ are lower and upper triangular matrices, respectively. By summing $L$ and $U$, the dense transformation matrix $D = L + U$ captures critical feature patterns in the data through linear transformations within the low-rank subspace defined by $A$.

The forward pass during fine-tuning is expressed as:
\begin{equation}\label{eq5}
    h = W_0 x + \alpha/(r + \epsilon) \cdot B(L + U)Ax,
\end{equation}
where $W_0 \in \mathbb{R}^{d \times n}$ is the pre-trained weight matrix, $x \in \mathbb{R}^{n}$ is the input, and $\alpha$ is a scaling factor ensuring training stability as the rank $r$ evolves. Both $L$ and $U$ are initialized with Gaussian random values, while $B$ is initialized as a zero matrix. The initial rank $r$ is set to 1, satisfying $r \ll \min(n, d)$, and is incremented dynamically during training. During training, $\Delta W$ is scaled by $\alpha/(r + \epsilon)$, where $\epsilon$ is a small constant to improve numerical stability.

The low-rank matrices $A$ and $B$, as well as the triangular matrices $L$ and $U$, are expanded during training to accommodate new dimensions. For example, the new $A^{(t)}$ is augmented by adding \(\Delta r\) rows at the end, as shown in Equation \ref{eq6}:
\begin{equation}\label{eq6}
A^{(t)} = \begin{bmatrix} A^{(t-1)} \\ A_{\text{aug}} \end{bmatrix},
\end{equation}
where \(A^{(t)} \in \mathbb{R}^{(r + \Delta r) \times n}\), \(A^{(t-1)} \in \mathbb{R}^{r \times n}\), and \(A_{\text{aug}} \in \mathbb{R}^{\Delta r \times n}\) is initialized with Gaussian random values.

Similarly, the new $B^{(t)}$ is augmented by adding \(\Delta r\) columns at the end, as shown in Equation \ref{eq7}:
\begin{equation}\label{eq7}
\begin{aligned}
B^{(t)} = \begin{bmatrix} B^{(t-1)} & B_{aug} \end{bmatrix},
\end{aligned}
\end{equation}
where \(B^{(t)} \in \mathbb{R}^{d \times (r + \Delta r)}\), \(B^{(t-1)} \in \mathbb{R}^{d \times r}\), and \(B_{\text{aug}} \in \mathbb{R}^{d \times \Delta r}\).

The lower triangular matrix \( L \) is augmented by adding \(\Delta r\) rows at the end to construct a new lower triangular matrix, as shown in Equation \ref{eq8}:
\begin{equation}\label{eq8}
\begin{aligned}
& L^{(t)} = \begin{bmatrix} L^{(t-1)} & 0 \\ L_{down} & L_{aug} \end{bmatrix},
\end{aligned}
\end{equation}
where \( L^{(t)} \in \mathbb{R}^{(r + \Delta r) \times (r + \Delta r)} \) is the augmented lower triangular matrix at time step \( t \), \( L^{(t-1)} \in \mathbb{R}^{r \times r} \) is the pre-existing lower triangular matrix from the previous time step \( t-1 \), \( L_{down} \in \mathbb{R}^{\Delta r \times r} \) is a dense matrix added to the bottom of \( L^{(t-1)} \), and \( L_{aug} \in \mathbb{R}^{\Delta r \times \Delta r} \) is the lower triangular matrix added to the bottom-right of the expanded matrix. Here, \( L_{down} \) is a dense matrix, meaning it is not constrained to a triangular structure, while \( L_{aug} \) retains the triangular form necessary for the structure of \( L^{(t)} \).

The upper triangular matrix \( U \) is augmented by adding \(\Delta r\) columns at the end to construct a new upper triangular matrix, as shown in Equation \ref{eq9}:
\begin{equation}\label{eq9}
\begin{aligned}
& U^{(t)} = \begin{bmatrix} U^{(t-1)} & U_{up} \\ 0 & U_{aug} \end{bmatrix},
\end{aligned}
\end{equation}
where \( U^{(t)} \in \mathbb{R}^{(r + \Delta r) \times (r + \Delta r)} \) is the augmented upper triangular matrix at time step \( t \), \( U^{(t-1)} \in \mathbb{R}^{r \times r} \) is the pre-existing upper triangular matrix from the previous time step \( t-1 \), \( U_{up} \in \mathbb{R}^{r \times \Delta r} \) is a dense matrix added to the right of \( U^{(t-1)} \), and \( U_{aug} \in \mathbb{R}^{\Delta r \times \Delta r} \) is the upper triangular matrix added to the bottom-right of the expanded matrix. Similarly, \( U_{up} \) is a dense matrix, and \( U_{aug} \) retains the upper triangular structure to preserve the triangular form of \( U^{(t)} \).

The linear combination of a lower triangular matrix \( L \) and an upper triangular matrix \( U \) allows for the simultaneous expansion of the parameters of the square dense transformation matrix \( D = L + U \) in both row and column directions, thereby achieving high parameter utilization and computational efficiency. This bidirectional parameter expansion approach enhances the scalability and stability of the incremental matrix while maintaining its low-rank characteristics.

Orthogonality Constraint. By constraining the low-rank matrices \(A\) and \(B\) to be orthogonal, parameter redundancy is effectively reduced, numerical stability is enhanced, feature diversity is preserved, gradient computations are simplified, and theoretical optimality and interpretability are provided. This is expressed in Equation \ref{eq10}~\cite{zhang2023adaptive}:
\begin{equation}\label{eq10}
\begin{aligned}
R(A,B) = \|A^T A - I\|_F^2 + \|B^T B - I\|_F^2.
\end{aligned}
\end{equation}

\subsection{Efficient Parameter Importance Evaluation}
The dense transformation matrix \(D = L + U \in \mathbb{R}^{r \times r}\) is employed to capture the critical features of the data within the low-rank subspace. We posit that the magnitude of its variation reflects the degree of change in the overall incremental matrix. Additionally, the Frobenius norm quantifies the overall magnitude of the matrix elements, that is, the ``energy" of the matrix. Therefore, the importance of an incremental matrix $\Delta W_m$ is determined by tracking changes in the normalized Frobenius norm of its dense transformation matrix $D$ over the training steps. The importance score $S_m^{(t)}$ for the $m$-th incremental matrix at time step $t$ is defined as:
\begin{equation}\label{eq11}
    S_m^{(t)} =  \frac{\| L^{(t)}_m+U^{(t)}_m \|_F}{r^{(t)}_m}  -  \frac{\| L^{(t-1)}_m+U^{(t-1)}_m \|_F}{r^{(t-1)}_m},
\end{equation}
where $\| \cdot \|_F$ denotes the Frobenius norm, and $r_m$ is the rank of $\Delta W_m$. This metric captures the relative contribution of $\Delta W_i$ to model adaptation.

Compared to AdaLoRA and IncreLoRA, our method significantly reduces the computational complexity of importance evaluation from \( \mathcal{O}((rn + dr)MT) \) to \( \mathcal{O}(r^2MT) \), where \( M \) denotes the number of incremental matrices, \( T \) represents the total number of iterations, and $r$, the rank, satisfies \( r \ll \min(n, d) \). This reduction translates to a substantial decrease in computational overhead during the model adaptation process for downstream tasks.

Moreover, the rank increment operation for the incremental matrices in our method is executed periodically, every few steps, rather than at every training step. This periodic update mechanism eliminates the need to compute importance scores based on gradient information at each step, a requirement in AdaLoRA and IncreLoRA that introduces significant computational burdens. For example, in the natural language understanding experiments detailed in Section~\ref{sec:nlu_label}, the MNLI task from the GLUE benchmark~\cite{wang2018glue} employs a rank increment interval of 1,000 steps. By avoiding the need for frequent importance evaluations, TriAdaptLoRA circumvents an additional computational overhead equivalent to 1,000 evaluations per interval. As a result, the overall computational cost of TriAdaptLoRA remains minimal, making it an efficient solution for rank adjustment in large-scale model adaptation.

\subsection{Adaptive Rank Growth with Dynamic Thresholds}

To efficiently allocate model parameters, we propose an adaptive rank growth algorithm based on dynamic thresholds. At each training step \( t \), the number of incremental matrices eligible for rank expansion, \( k^{(t)} \), is dynamically determined by a thresholding scheme, which can be linear or nonlinear. The rank budget \( \mathfrak{R}^{(t)} \) is updated iteratively as:
\begin{equation}
\mathfrak{R}^{(t)} = \begin{cases}\mathfrak{R}^{(0)} - r \cdot M & \text { if } t = 1 \\ \mathfrak{R}^{(t-1)} - \Delta r \cdot k^{(t-1)} & \text { if } t > 1\end{cases},
\end{equation}
where \(\mathfrak{R}^{0} = r^{ref} \cdot M\) is the initial total rank budget, \(r^{ref}\) is the reference rank, $r$ is the initial rank of the low-rank matrix, and \(\Delta r\) is the rank increment per matrix. To ensure at least one matrix is always selected, we impose a lower bound \( k^{(t)} = \max(k^{(t)}, 1) \).

\paragraph{Linear Threshold} 
In the linear strategy, rank growth accelerates early to enhance model capacity and stabilizes in later stages for fine-tuning:
\begin{equation}\label{eqlinear}
k^{(t)} = \lceil \mathfrak{R}^{(t)} \cdot \alpha^{(t)} \rceil, \quad \alpha^{(t)} = \frac{t - t_0}{T - t_0}.
\end{equation}
Here, \( \mathfrak{R}^{(t)} \) represents the remaining rank budget at step \( t \), and \( \alpha^{(t)} \) determines the fraction of the budget to allocate dynamically.

\paragraph{Nonlinear Threshold} 
The nonlinear strategy delays rank growth initially to prevent overfitting, then accelerates it in later stages to enhance model expressiveness:
\begin{equation}\label{eqnonlinear}
k^{(t)} = \lceil e^{\alpha^{(t)} \cdot \log \mathfrak{R}^{(t)}} \rceil.
\end{equation}
This mode is particularly suitable for tasks requiring gradual early-stage adjustments and rapid expansion in later stages.

\paragraph{Rank Growth}
At each step, the importance threshold \( S^{\theta}(t) \) is:
\[
S^{\theta}(t) = \min\{S(m) | m \in \text{top-}k^{(t)}\}.
\]
Matrices with scores \( S^{(t)}_m \geq S^{\theta}(t) \) are selected for rank increment, and new rows and columns in matrices \( A \), \( B \), \( L \), and \( U \) are initialized with Gaussian random values.

The adaptive rank growth algorithm dynamically determines the number of incremental matrices involved in rank growth during the iterative process, based on budget allocation. This effectively mitigates the inefficiency of parameter usage inherent in traditional fixed-threshold approaches, ensuring a balance between parameter efficiency and model expressiveness.

The proposed adaptive rank growth algorithm significantly improves over fixed-threshold approaches. By incorporating both linear and nonlinear strategies, it dynamically adjusts parameter allocation to better adapt to task complexity. The method is particularly suitable for scenarios requiring early-stage performance boosts or late-stage fine-grained adjustments.

\subsection{Analysis of the Role of the Transformation Matrix}

Using gradient information, we further analyze the importance of the dense 
 transformation matrix \( D = L + U \) within the incremental matrices. Assuming the loss function is \( \mathcal{L} \), the gradient of \( \mathcal{L} \) with respect to \( \Delta W \) is given by Equation \ref{eq15}:
\begin{equation}\label{eq15}
\begin{aligned}
& \frac{\partial \mathcal{L}}{\partial \Delta W} = \frac{\partial \mathcal{L}}{\partial h} \cdot \frac{\partial h}{\partial \Delta W} \\
& \phantom{\frac{\partial \mathcal{L}}{\partial \Delta W}} = \frac{\partial \mathcal{L}}{\partial h} \cdot (I \otimes x)_{reorder(3,(1,2))},
\end{aligned}
\end{equation}
where \( I \) denotes the identity matrix, $reorder$ refers to the reordering of dimensions, and \( \otimes \) represents the tensor product. Utilizing the chain rule of derivatives, we can further derive the gradient information of the loss function \( \mathcal{L} \) with respect to the matrices \( B \), \( L \), \( U \), and \( A \) as shown in Equation \ref{eq16}:
\begin{equation}\label{eq16}
\begin{aligned}
& \frac{\partial \mathcal{L}}{\partial B} = \frac{\partial \mathcal{L}}{\partial h} \cdot \frac{\partial h}{\partial \Delta W} \cdot \frac{\partial B(L + U) A}{\partial B} \\
& \phantom{\frac{\partial \mathcal{L}}{\partial B}} = \frac{\partial \mathcal{L}}{\partial h} \cdot (I \otimes ((L + U)Ax))_{reorder(3,(1,2))} \\
& \frac{\partial \mathcal{L}}{\partial L} = \frac{\partial \mathcal{L}}{\partial h} \cdot \frac{\partial h}{\partial \Delta W} \cdot \frac{\partial B(L + U) A}{\partial L} \\
& \phantom{\frac{\partial \mathcal{L}}{\partial L}} = \frac{\partial \mathcal{L}}{\partial h} \cdot (B^T \otimes (Ax))_{reorder(3,(1,2))} \\
& \frac{\partial \mathcal{L}}{\partial U} = \frac{\partial \mathcal{L}}{\partial h} \cdot \frac{\partial h}{\partial \Delta W} \cdot \frac{\partial B(L + U) A}{\partial U} \\
& \phantom{\frac{\partial \mathcal{L}}{\partial U}}= \frac{\partial \mathcal{L}}{\partial h} \cdot (B^T \otimes (Ax))_{reorder(3,(1,2))} \\
& \frac{\partial \mathcal{L}}{\partial A} = \frac{\partial \mathcal{L}}{\partial h} \cdot \frac{\partial h}{\partial \Delta W} \cdot \frac{\partial B(L + U) A}{\partial A} \\
& \phantom{\frac{\partial \mathcal{L}}{\partial A}} = \frac{\partial \mathcal{L}}{\partial h} \cdot (((U^T + L^T)B^T) \otimes x)_{reorder(3,(1,2))}.
\end{aligned}
\end{equation}

During the backward propagation of model gradients, the dense transformation matrix \( D = L + U \) plays a crucial role in the parameter update process. Its variations directly affect the updates of the low-rank matrices \( A \) and \( B \), which in turn regulate the optimization of \( L \) and \( U \). Specifically, the gradient update of the low-rank matrix \( A \) involves tensor product operations. For example, the gradient is computed based on the combination of input \( x \) and the matrix \( (U^\top + L^\top)B^\top \). These multi-layer tensor operations couple the features of \( x \) with the matrices \( B \), \( L \), and \( U \), thereby extending the gradient propagation path. Since the gradient updates are confined within the subspace defined by the transformation matrix \( U^\top + L^\top \), this mechanism implicitly regularizes the parameter updates, effectively suppressing drastic parameter changes and reducing the risk of overfitting. Therefore, monitoring the changes in the Frobenius norm of the transformation matrix to evaluate the importance of incremental matrices is justified.

\begin{algorithm}[h]
\caption{TriAdaptLoRA Training Procedure}
\label{alg:trilora}
\begin{algorithmic}[1]
\State \textbf{Input:} Dataset $\mathcal{G}$, Total Steps $T$, Warmup Steps $t_0$, Learning Rate $\eta$, Total Rank Budget $\mathfrak{R}^0$
\State \textbf{Initialization:} $A, B, L, U$ with $r=1$
\For{$t = 1$ to $T$}
    \If{$\mathfrak{R}^{(t-1)} > 0$}
        \State Compute the importance scores $S_m^{(t)}$ for each $\Delta W_m$ using Equation \ref{eq11}
        \State Compute the threshold \( k^{(t)} \) using the linear threshold model (Equation \ref{eqlinear}) or the nonlinear threshold model (Equation \ref{eqnonlinear})
        \State Increment the ranks of matrices \( A \), \( B \), \( L \), and \( U \) within the top-$k^{(t)}$ incremental matrices using Equations \ref{eq6}, \ref{eq7}, \ref{eq8}, and \ref{eq9}, respectively
    \EndIf
    \State Update $A, B, L, U$ via gradient descent
\EndFor
\State \textbf{Output:} Fine-tuned parameters $A, B, L, U$
\end{algorithmic}
\end{algorithm}

\section{Experiments}

This experiment aims to validate the effectiveness of the TriAdaptLoRA method in natural language processing tasks. Specifically, the experiment encompasses nine tasks across two major categories: Natural Language Understanding (NLU) and Natural Language Generation (NLG). Fine-tuning was performed using methods such as TriAdaptLoRA, AdaLoRA, and IncreLoRA based on the pre-trained model DeBERTaV3-base~\cite{he2023debertav}. To comprehensively assess the effectiveness of the TriAdaptLoRA method, the experimental results of TriAdaptLoRA will also be compared with existing fine-tuning methods, including Full Fine-Tuning, Adaptive Tuning, BitFit, and LoRA.

\subsection{Baseline}

The experiment selects several representative fine-tuning methods from recent years for comparison to comprehensively evaluate the effectiveness of TriAdaptLoRA in adapting to downstream tasks. The following provides a detailed description of these fine-tuning methods:

\begin{itemize}
\item \textbf{Full Fine-Tuning}: Updates all pre-trained model parameters, achieving high performance but with significant resource overhead~\cite{qiu2020pre, raffel2020exploring}.
\item \textbf{Adapter Tuning}: Introduces adapter modules, reducing trainable parameters while maintaining performance, though increasing model complexity and inference latency~\cite{rebuffi2017learning, houlsby2019parameter, he2022towards}.
\item \textbf{BitFit}: Adjusts only bias parameters, substantially reducing trainable parameters and performing well across tasks~\cite{zaken2021bitfit}.
\item \textbf{LoRA}: A fundamental parameter-efficient fine-tuning method, from which numerous optimization techniques have been derived from various perspectives~\cite{hu2022lora}.
\item \textbf{AdaLoRA}: Enhances performance by pruning and optimizing the rank of incremental matrices~\cite{zhang2023adaptive}.
\item \textbf{IncreLoRA}: Improves performance by progressively increasing the rank of incremental matrices~\cite{Zhang2023IncreLoRAIP}.
\end{itemize}

\subsection{Natural Language Understanding Tasks}
\label{sec:nlu_label}
\subsubsection{Base Model and Tasks}

This experiment employs DeBERTaV3-base~\cite{he2023debertav}, an advanced pre-trained language model from Microsoft Research.

To validate the effectiveness of the TriAdaptLoRA method in natural language understanding tasks, the experiment utilizes the GLUE benchmark (General Language Understanding Evaluation)~\cite{wang2018glue}, encompassing the following eight tasks:

\begin{itemize}
\item \textbf{CoLA} (Corpus of Linguistic Acceptability): Assesses the grammatical acceptability of English sentences, utilizing the Matthews correlation coefficient as the evaluation metric ~\cite{warstadt2019neural,matthews1975comparison}.
\item \textbf{MNLI} (Multi-Genre Natural Language Inference): Determines the textual entailment relationship between premise and hypothesis sentences across multiple domains~\cite{williams2017broad}.
\item \textbf{MRPC} (Microsoft Research Paraphrase Corpus): Assesses paraphrase relationships between sentence pairs, with an imbalanced data distribution (68\% positive samples)~\cite{dolan2005automatically}.
\item \textbf{QNLI} (Question Natural Language Inference): Based on the Stanford Question Answering Dataset, it determines whether a statement contains the answer to a question~\cite{wang2018glue,rajpurkar2016squad}.
\item \textbf{QQP} (Quora Question Pairs): Evaluates the semantic similarity of two question sentences, with an imbalanced data distribution (63\% negative samples)~\cite{wang2018glue}.
\item \textbf{RTE} (Recognizing Textual Entailment): A small-scale textual entailment dataset that integrates results from multiple RTE challenges~\cite{dagan2005pascal,haim2006second,giampiccolo2007third,bentivogli2009fifth}.
\item \textbf{SST-2} (Stanford Sentiment Treebank): A sentence-level sentiment classification task based on movie reviews (positive/negative)~\cite{socher2013recursive}.
\item \textbf{STS-B} (Semantic Textual Similarity Benchmark): Measures the semantic similarity of sentence pairs using Pearson and Spearman correlation coefficients~\cite{cer2017semeval}.
\end{itemize}

Detailed task statistics for the GLUE benchmark are provided in Appendix \ref{glue_statistics}.

\subsubsection{Experimental Details}
\label{sec:nlu_detail_label}

The experiments use the DeBERTaV3-base pre-trained model as the backbone to apply and evaluate the TriAdaptLoRA, AdaLoRA, and IncreLoRA methods across the eight GLUE benchmark tasks.
The training and testing processes adhere to the following settings:
\begin{itemize}
\item Optimizer: AdamW is selected as the optimizer, with a linear learning rate decay strategy employed.
\item Hardware: The GPU processor used is NVIDIA A100-PCIE-40GB.
\item Hyperparameters: The hyperparameters, including learning rate, reference rank ($r^{ref}$) size, batch size, and number of training epochs, are consistent with the settings of Zhang et al. (2023)~\cite{Zhang2023IncreLoRAIP}. Detailed hyperparameter configurations are available in Appendix \ref{NLU_super_para}.
\item Configurations: TriAdaptLoRA, LoRA, AdaLoRA and IncreLoRA are applied to all linear weight matrices. Specifically, these include the query ($W_q$), key ($W_k$), and value ($W_v$) projection matrices, the intermediate projection matrix ($W_m$), the attention output projection matrix ($W_a$), and the layer output projection matrix ($W_o$).
\item For each task, the reported results are the averages and standard deviations obtained from training and testing with three different random seeds.
\end{itemize}

\subsubsection{Experimental Results}
\label{sec:nlu_results_label}
This study examines the impact of two adaptive rank growth threshold modes for incremental matrices in the TriAdaptLoRA method on natural language understanding task performance. Using the DeBERTaV3-base model, we conducted comparative experiments on the eight GLUE benchmark tasks employing TriAdaptLoRA, LoRA, AdaLoRA, and IncreLoRA methods.

\begin{table*}[htbp]
    \caption{DeBERTaV3-base fine-tuned on GLUE benchmark tasks. Metrics include accuracy (MNLI, QNLI), Matthews correlation (CoLA), and Pearson correlation (STS-B).}
    \label{Table1}
    \centering
    \resizebox{\textwidth}{!}{
    \begin{tabular}{m{3.3cm}|cccccccc|c}
        \hline \rule{0pt}{2.5ex}\textbf{Model\&Method (DeBERTaV3-base)} & \textbf{MNLI} & \textbf{SST-2} & \textbf{CoLA} & \textbf{QQP} & \textbf{QNLI} & \textbf{RTE} & \textbf{MRPC} & \textbf{STS-B} & \textbf{Avg.} \\
        \hline
        \rule{0pt}{2.5ex}Full Fine-tuning & $89.9$ & $95.63$ & $69.19$ & $\mathbf{92.4}$ & $94.03$ & $83.75$ & $89.46$ & $91.6$ & $88.24$ \\
        BitFit & $89.37$ & $94.84$ & $66.96$ & $88.41$ & $92.24$ & $78.7$ & $87.75$ & $91.35$ & $86.20$ \\
        HAdapter & $90.13$ & $95.53$ & $68.64$ & $91.91$ & $94.11$ & $84.48$ & $89.95$ & $91.48$ & $88.12$ \\
        LoRA & $89.61_{\pm .1}$ & $93.54_{\pm .2}$ & $65.64_{\pm 1.0}$ & $92.19_{\pm .0}$ & $92.55_{\pm .1}$ & $82.07_{\pm 1.1}$ & $88.64_{\pm .2}$ & $91.54_{\pm 0}$ & $86.97$ \\
        AdaLoRA & $\mathbf{90.66}_{\pm 0}$ & $\mathbf{95.75}_{\pm .2}$ & $70.21_{\pm 1.0}$ & $92.23_{\pm 0}$ & $\mathbf{94.51}_{\pm .2}$ & $86.88_{\pm .6}$ & $90.03_{\pm .4}$ & $91.69_{\pm .1}$ & $88.99$ \\
        IncreLoRA & $90.62_{\pm 0}$ & $95.72_{\pm .3}$ & $70.2_{\pm .1}$ & $91.91_{\pm 0}$ & $94.36_{\pm .2}$ & $86.88_{\pm .4}$ & $90.11_{\pm .4}$ & $91.38_{\pm .2}$ & $88.9$ \\
        TriAdaptLoRA (Fixed-$k$) & $90.5_{\pm 0}$ & $95.45_{\pm .4}$ & $\mathbf{71.87}_{\pm .4}$ & $91.87_{\pm 0}$ & $94.43_{\pm 0}$ & $87.36_{\pm 1.1}$ & $89.46_{\pm 1.2}$ & $91.65_{\pm .1}$ & $89.07$ \\
        TriAdaptLoRA & $90.64_{\pm .1}$ & $95.68_{\pm .2}$ & $71.6_{\pm 0}$ & $92.09_{\pm 0}$ & $94.37_{\pm 0}$ & $\mathbf{87.84}_{\pm .6}$ & $\mathbf{90.77}_{\pm .3}$ & $\mathbf{91.79}_{\pm .1}$ & $\mathbf{89.34}$ \\
        \hline
        \rule{0pt}{2.5ex}TriAdaptLoRA (Linear) & $90.63_{\pm .2}$ & $95.3_{\pm .1}$ & $71.6_{\pm .4}$ & $92.09_{\pm 0}$ & $94.37_{\pm 0}$ & $\mathbf{87.84}_{\pm .6}$ & $\mathbf{90.77}_{\pm .3}$ & $91.68_{\pm .1}$ & $89.28$ \\
        TriAdaptLoRA (Non-Linear) & $90.64_{\pm .1}$ & $95.68_{\pm .2}$ & $71.6_{\pm 0}$ & $91.88_{\pm 0}$ & $94.25_{\pm .1}$ & $86.76_{\pm .1}$ & $90.2_{\pm 0}$ & $\mathbf{91.79}_{\pm .1}$ & $89.1$ \\
        \hline
    \end{tabular}}
\end{table*}

The results in Table \ref{Table1} demonstrate that TriAdaptLoRA significantly outperforms IncreLoRA, particularly by substantially reducing the computational overhead of the rank adjustment process while still achieving an overall performance improvement of approximately 0.44\% on GLUE benchmark tasks. The specific task performances are as follows:

\begin{itemize}
\item  Linear Threshold Mode: TriAdaptLoRA enhances performance by 1.4\% ($\sigma$=0.4\%) on CoLA, 0.96\% ($\sigma$=0.6\%) on RTE, and 0.66\% ($\sigma$=0.3\%) on MRPC, demonstrating significant advantages on small-scale datasets. The linear threshold mode achieves steady performance improvements across most tasks, reflecting high adaptability. In all cases, $\sigma$ represents the standard deviation.
\item Non-Linear Threshold Mode: TriAdaptLoRA exhibits better adaptability on tasks with larger data scales. For instance, it achieves superior results on MNLI and SST-2 tasks and improves STS-B performance by 0.41\% ($\sigma$=0.1\%). These results indicate that the non-linear threshold mode complements the linear growth mode in complex tasks.
\end{itemize}

\begin{figure*}[htb]
  \centering
  \includegraphics [width=18cm]{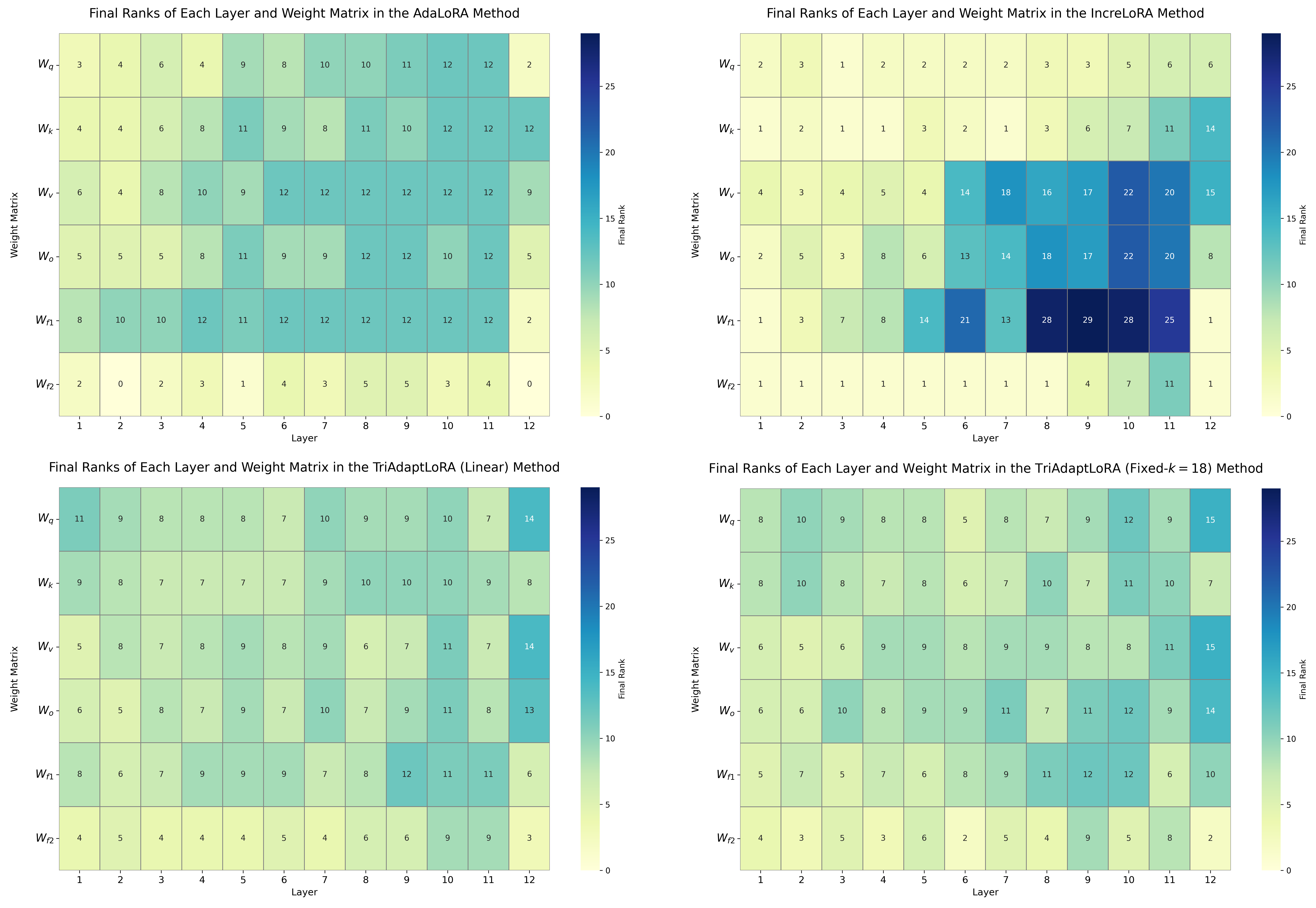}
  \caption{Heatmaps illustrate the final rank distributions. These distributions are shown for each layer and weight matrix of the TriAdaptLoRA, AdaLoRA, and IncreLoRA methods on the MNLI task.}
  \label{P2}
\end{figure*}

Furthermore, Figure \ref{P2} illustrates the rank distributions across various layers and weight matrices for the TriAdaptLoRA, AdaLoRA, and IncreLoRA methods after training completion. 
The results indicate that the rank distributions of TriAdaptLoRA and AdaLoRA are similar, with the majority of ranks concentrated in the higher layers' weight matrices. However, AdaLoRA imposes an upper limit on ranks and exhibits insufficient continuity and uniformity. In contrast, TriAdaptLoRA demonstrates a more uniform, continuous, and consistent rank distribution under both Linear and Fixed-$k$ modes. 
This characteristic enhances the model's generalization ability and stability while achieving higher parameter efficiency. Conversely, IncreLoRA's rank distribution is predominantly concentrated in specific regions, and this dominant parameter allocation approach may increase the risk of overfitting, thereby detrimentally affecting the model's generalization performance. These findings further validate the effectiveness of the parameter importance evaluation method employed in TriAdaptLoRA, indicating that TriAdaptLoRA can efficiently allocate parameters to critical modules, thereby optimizing overall model performance.

\subsubsection{Ablation Studies}

To further validate the effectiveness of the TriAdaptLoRA method, ablation experiments are conducted and analyzed in three aspects:

\textbf{Orthogonal Constraint Analysis}: This experiment explores the impact of introducing orthogonal constraints in the TriAdaptLoRA method on model performance. Specifically, it compares model performance with and without orthogonal constraints in low-rank matrices. The results in Table \ref{Table2} show that applying orthogonal constraints in low-rank matrices generally enhances model performance. For example, under the non-linear threshold mode, imposing orthogonal constraints in low-rank matrices leads to better performance on MNLI, SST-2, and MRPC tasks compared to scenarios without orthogonal constraints.

However, in CoLA and RTE tasks, orthogonal constraints result in decreased performance, indicating that the effectiveness of orthogonal constraints varies across different tasks. Therefore, the application of orthogonal constraints should be flexibly adjusted based on task characteristics to achieve optimal performance.

\begin{table*}[htbp]
    \caption{Impact of orthogonal constraints on TriAdaptLoRA performance across GLUE benchmark tasks.}
    \label{Table2}
    \centering
    \resizebox{\textwidth}{!}{
    \begin{tabular}{m{3.3cm}|cccccccc|c}
        \hline \rule{0pt}{4.5ex}\textbf{Model\&Method (DeBERTaV3-base)} & \textbf{MNLI} & \textbf{SST-2} & \textbf{CoLA} & \textbf{QQP} & \textbf{QNLI} & \textbf{RTE} & \textbf{MRPC} & \textbf{STS-B} & \textbf{Avg.} \\
        \hline
        \rule{0pt}{2.5ex}TriAdaptLoRA (Non-Linear) (Orthogonal) & $90.64_{\pm .1}$ & $95.68_{\pm .2}$ & $71.6_{\pm 0}$ & $91.88_{\pm 0}$ & $94.25_{\pm .1}$ & $86.76_{\pm .1}$ & $90.2_{\pm 0}$ & $91.79_{\pm .1}$ & $89.1$ \\
        \hline
        \rule{0pt}{2.5ex}TriAdaptLoRA (Non-Linear) (Non-Orthogonal) & $90.24 \pm 0$ & $95.52 \pm 0$ & $\mathbf{71.72} \pm .6$ & $91.89 \pm 0$ & $94.11 \pm .1$ & $\mathbf{86.88} \pm .7$ & $89.87 \pm .3$ & $91.78 \pm .1$ & $89$ \\
        \hline
    \end{tabular}}
\end{table*}

\textbf{Normalization Analysis of Frobenius Norm}: To investigate the effect of different Frobenius norm normalization strategies in TriAdaptLoRA on model performance, three approaches were designed: (1) Frobenius norm divided by rank $r_m$; (2) Frobenius norm divided by rank $\sqrt{r_m}$; and (3) no normalization (direct use of Frobenius norm). As shown in Table \ref{Table3}, the strategy of dividing the Frobenius norm by rank $r_m$ significantly improves performance in most tasks, becoming the default scheme for the main experiments (Sections \ref{sec:nlu_results_label} and \ref{sec:nlg_results_label}). Additionally, the normalization scheme of dividing the Frobenius norm by rank $\sqrt{r_m}$ also shows certain advantages, albeit slightly inferior to the former. In contrast, the no-normalization scheme performs the worst, further confirming the importance of normalization for model performance.

The lack of normalization allows incremental matrices with larger ranks to dominate the training process, diminishing the optimization opportunities for smaller-rank incremental matrices and affecting the reasonable distribution of parameters and the model’s generalization ability. By using the Frobenius norm divided by rank $r_m$ normalization strategy, the differences in ranks among incremental matrices are balanced, enabling smaller-rank incremental matrices to also have opportunities for parameter expansion, thereby enhancing model performance.

\begin{table*}[htbp]
    \caption{TriAdaptLoRA performance with different Frobenius norm normalization schemes across eight GLUE benchmark tasks.}
    \label{Table3}
    \centering
    \resizebox{\textwidth}{!}{
    \begin{tabular}{m{3.3cm}|cccccccc|c}
        \hline \rule{0pt}{4.5ex}\textbf{Model\&Method (DeBERTaV3-base)} & \textbf{MNLI} & \textbf{SST-2} & \textbf{CoLA} & \textbf{QQP} & \textbf{QNLI} & \textbf{RTE} & \textbf{MRPC} & \textbf{STS-B} & \textbf{Avg.} \\
        \hline
        \rule{0pt}{2.5ex}TriAdaptLoRA (Non-Linear) (F-norm divided by $r_m$) & $90.64_{\pm .1}$ & $95.68_{\pm .2}$ & $71.6_{\pm 0}$ & $91.88_{\pm 0}$ & $94.25_{\pm .1}$ & $86.76_{\pm .1}$ & $90.2_{\pm 0}$ & $91.79_{\pm .1}$ & $89.1$ \\   
        \hline 
        \rule{0pt}{2.5ex}TriAdaptLoRA (Non-Linear) (F-norm not divided by $r_m$) & $90.47 \pm 0.1$ & $95.37 \pm 0$ & $71.39 \pm 0.1$ & $91.64 \pm 0$ & $94.01 \pm 0.1$ & $86.76 \pm 0.6$ & $89.87 \pm 0.4$ & $91.43 \pm 0.1$ & $88.86$ \\
        \hline 
        \rule{0pt}{2.5ex}TriAdaptLoRA (Non-Linear) (F-norm divided by $\sqrt{r_m}$) & $90.5 \pm 0.1$ & $95.41 \pm 0$ & $71.36 \pm 0.2$ & $91.8 \pm 0$ & $94.46 \pm 0.1$ & $86.76 \pm 1.1$ & $89.38 \pm 0.2$ & $91.55 \pm 0$ & $88.9$ \\
        \hline
    \end{tabular}}
\end{table*}

\textbf{Fixed Rank Growth Threshold Mode Analysis}: As shown in Table \ref{Table1}, the fixed rank growth threshold mode (Fixed‑$k$) of TriAdaptLoRA outperforms IncreLoRA and AdaLoRA overall on the GLUE benchmark tasks. Notably, IncreLoRA employs the same fixed rank growth threshold mode (Fixed‑$k$), demonstrating the effectiveness of the parameter expansion mechanism through triangular split and the parameter importance evaluation algorithm. However, compared to the adaptive rank growth threshold mode, the overall performance improvement is relatively modest, indicating that the adaptive rank growth threshold mode can further enhance parameter efficiency. Additionally, the fixed rank growth threshold mode is highly sensitive to the setting of the hyperparameter \( k \), necessitating tedious tuning processes for different tasks, which increases the adaptation cost and complexity in multi-task scenarios. In contrast, adaptively adjusted linear and nonlinear threshold modes exhibit better performance and adaptability in multi-task environments and do not rely on task-specific hyperparameter tuning.

\subsection{Natural Language Generation Tasks}

\subsubsection{Base Model and Tasks}

This experiment employs the DeBERTaV3-base pre-trained model released by Microsoft and evaluates the fine-tuning performance based on the \textbf{SQuAD 2.0} (Stanford Question Answering Dataset 2.0)~\cite{rajpurkar-etal-2018-know} task. SQuAD 2.0 introduces 53,775 unanswerable questions in addition to the original SQuAD 1.1~\cite{rajpurkar2016squad}, aiming to assess the model's ability to identify questions that cannot be answered. 

\subsubsection{Experimental Details}

This experiment conducts experiments on the SQuAD 2.0 task by applying TriAdaptLoRA and IncreLoRA methods on the DeBERTaV3-base pre-trained model. The following settings are used during training and testing:

\begin{itemize}
\item Optimizer: AdamW is selected as the optimizer, with a linear learning rate decay strategy employed.
\item Hardware: The GPU processor used is NVIDIA A100-PCIE-40GB.
\item Hyperparameters: The hyperparameters, including learning rate, reference rank ($r^{ref}$) size, batch size, and number of training epochs, are consistent with the settings of Zhang et al. (2023)~\cite{Zhang2023IncreLoRAIP}. Detailed hyperparameter configurations are available in Appendix \ref{NLG_super_para}.
\item Configurations: TriAdaptLoRA and IncreLoRA are applied to all linear weight matrices. Specifically, these include the query ($W_q$), key ($W_k$), and value ($W_v$) projection matrices, the intermediate projection matrix ($W_m$), the attention output projection matrix ($W_a$), and the layer output projection matrix ($W_o$).
\end{itemize}

\subsubsection{Experimental Results}
\label{sec:nlg_results_label}

In this experiment, we investigate the impact of three different rank growth threshold modes in the TriAdaptLoRA method on natural language generation tasks. Based on the DeBERTaV3-base pre-trained model, we evaluated TriAdaptLoRA, LoRA, IncreLoRA, HAdapter, and Full Fine-Tuning methods on the SQuAD 2.0 task. The experimental results are presented in Table \ref{Table4}.

\begin{table}[htbp]
	\caption{Fine-tuning results of DeBERTaV3-base on the SQuAD 2.0 task.}
	\label{Table4}
    \centering
    \small
    \setlength{\tabcolsep}{3.6pt} 
    \begin{tabular}{l|ccc}
        \hline \rule{0pt}{2.5ex}\textbf{Method} & \textbf{SQuAD 2.0(EM)} & \textbf{SQuAD 2.0(F1)}\\
        \hline
        \rule{0pt}{2.5ex}Full Fine-Tuning & $85.4$ & $88.4$ \\
        HAdapter & $85.4$ & $88.3$ \\
        LoRA & $85.0$ & $88.0$ \\
        IncreLoRA & $85.56$ & $88.66$ \\
        TriAdaptLoRA (Linear) & $\mathbf{8 5.82}$ & $\mathbf{88.9}$ \\
        TriAdaptLoRA (Non-Linear) & $85.63$ & $88.68$ \\
        TriAdaptLoRA (Fixed-$k$) & $85.24$ & $88.21$ \\
        \hline
    \end{tabular}
\end{table}

From the experimental results in Table \ref{Table4}, it is evident that compared to IncreLoRA, the two adaptive rank growth threshold modes of TriAdaptLoRA both achieve performance improvements on the SQuAD 2.0 task. Specifically, the linear threshold mode increases the model's EM (Exact Match) and F1 scores by approximately 0.26\% and 0.24\%, respectively, outperforming other comparative methods. The nonlinear threshold mode also brings performance gains and surpasses the full fine-tuning method, although its improvement is slightly smaller compared to the linear threshold mode. Additionally, the fixed threshold mode results in a slight performance decline, further highlighting its limitations in adapting to downstream tasks. The experimental results validate the effectiveness of TriAdaptLoRA in natural language generation tasks, particularly demonstrating that the linear threshold mode can effectively enhance model performance.

\subsection{Experimental Analysis}

To further validate the effectiveness of the TriAdaptLoRA method, we will conduct additional experiments and detailed analyses from three aspects:

\subsubsection{Sensitivity to Initial Warm-up Steps}

\begin{figure}[htb]
  \centering
  \includegraphics [width=9cm]{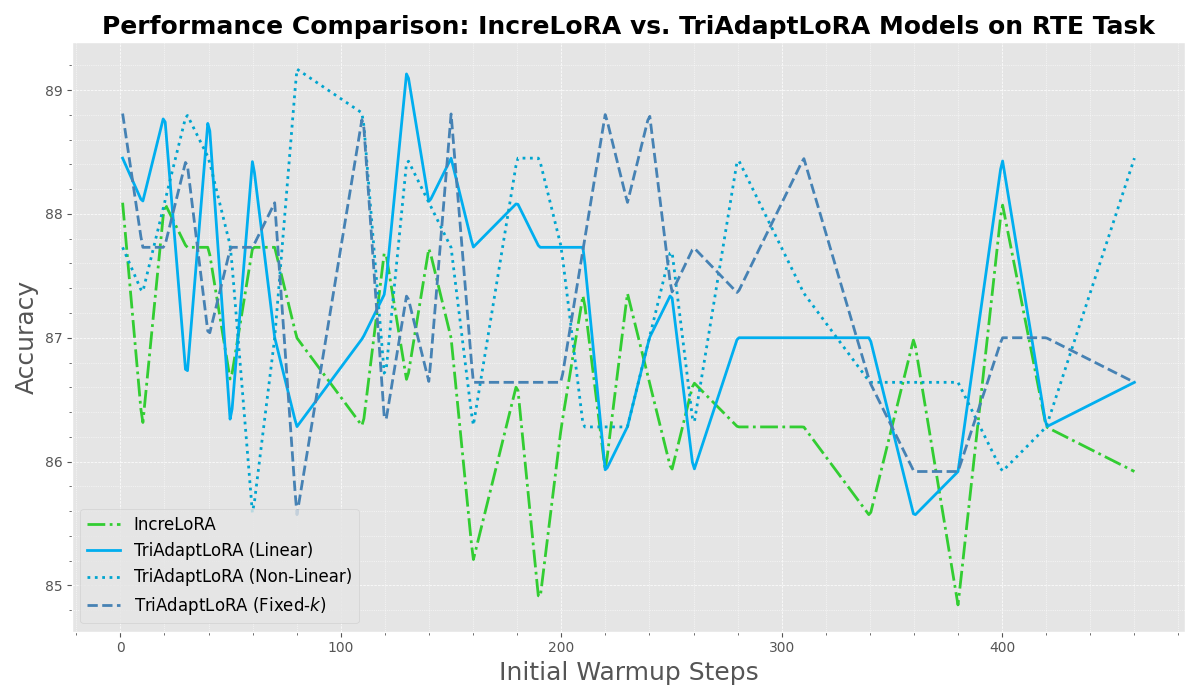}
  \caption{Experimental results on the RTE task with different initial warm-up steps for TriAdaptLoRA and IncreLoRA methods based on the DeBERTaV3-base model.}
  \label{P3}
\end{figure}

This experiment is based on the DeBERTaV3-base model and compares the performance of TriAdaptLoRA and IncreLoRA methods under different initial warm-up steps settings to assess the impact of initial warm-up steps on model performance and stability. The experiment is conducted on the RTE task, keeping all other hyperparameters and random seeds consistent. Detailed experimental settings are provided in Section \ref{sec:nlu_detail_label}. Figure \ref{P3} illustrates the effect of different initial warm-up steps on model performance.

Key conclusions are as follows: 
\begin{itemize}
\item Initial Warm-up Steps Significantly Affect Model Performance: Based on the performance of TriAdaptLoRA and IncreLoRA, a reasonable number of warm-up steps can effectively improve model performance, possibly because it helps the model gradually adapt to task characteristics. However, too few or too many warm-up steps may lead to performance fluctuations, indicating that the initial warm-up steps need to be optimized according to task requirements.  
\item Relationship Between Performance and Initial Warm-up Steps: As the number of initial warm-up steps increases, the overall performance of both TriAdaptLoRA and IncreLoRA tends to decline. However, TriAdaptLoRA demonstrates higher stability, indicating its advantage in parameter adjustment and task adaptability. In contrast, IncreLoRA is more sensitive to initial warm-up steps, exhibiting larger performance fluctuations.  
\item Application Potential: TriAdaptLoRA consistently shows good performance and stability under different initial warm-up settings, validating its adaptability and robustness in fine-tuning pre-trained models. This characteristic gives TriAdaptLoRA practical potential in complex task scenarios.
\end{itemize}

\subsubsection{Sensitivity to Rank Update Interval}

\begin{figure}[htb]
  \centering
  \includegraphics [width=9cm]{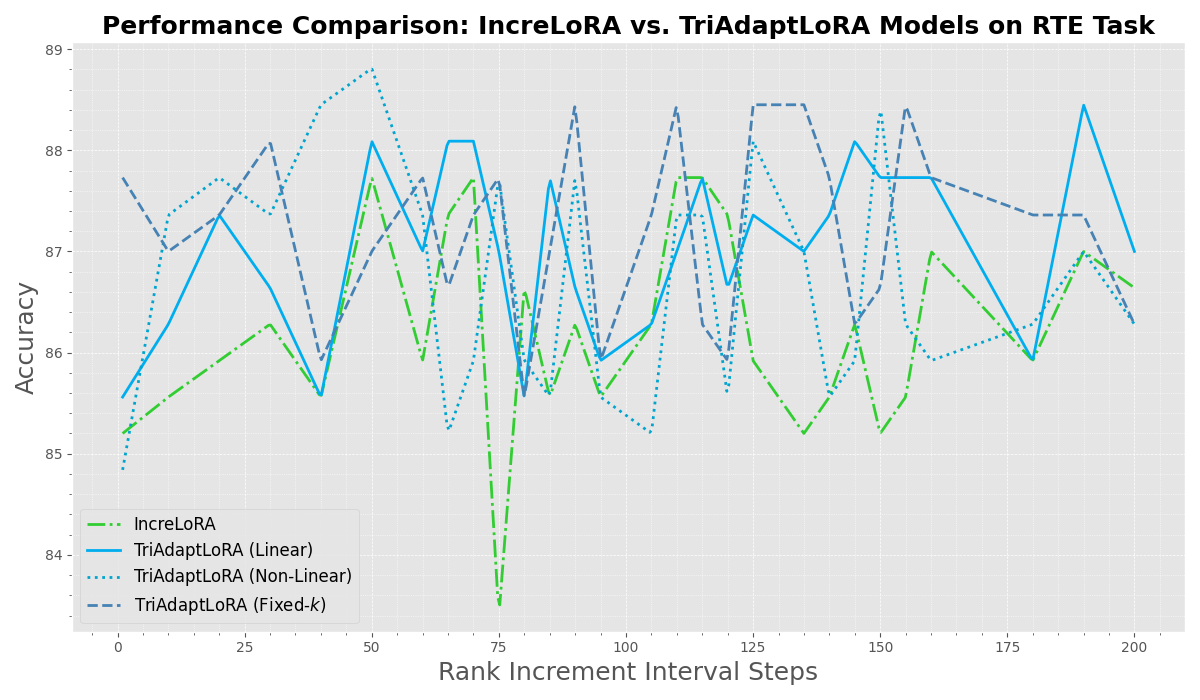}
  \caption{Experimental results on the RTE task with different rank update intervals for TriAdaptLoRA and IncreLoRA methods based on the DeBERTaV3-base model.}
  \label{P4}
\end{figure}

This experiment aims to assess the impact of different rank update intervals on the model performance and stability of TriAdaptLoRA and IncreLoRA methods. The experiment is based on the DeBERTaV3-base pre-trained model on the RTE task, with all other hyperparameters and random seeds kept consistent. Experimental settings are provided in Section \ref{sec:nlu_detail_label}. Figure \ref{P4} shows the comparison of model performance under different rank update intervals.

Key conclusions are as follows: 
\begin{itemize}
\item Performance: TriAdaptLoRA outperforms IncreLoRA in most rank update interval settings, demonstrating its effectiveness and superiority in optimizing model performance.  
\item Impact of Rank Update Interval on Performance: As the rank update interval increases, both TriAdaptLoRA and IncreLoRA show an initial improvement in performance, which then stabilizes. The best performance is achieved when the rank update interval is in the optimal range, indicating that a reasonable interval is crucial for improving model performance. Too many or too few rank updates may limit the model’s potential for utilizing rank growth, thereby affecting overall performance.  
\item Stability and Efficiency: TriAdaptLoRA exhibits higher stability across different rank update intervals, indicating its lower sensitivity to the rank update interval. As the interval increases, model complexity and computational overhead gradually decrease, reflecting TriAdaptLoRA’s greater robustness and resource efficiency. This characteristic is especially important for reducing energy consumption.
\end{itemize}

\subsubsection{Sensitivity to Reference Rank}

\begin{figure}[htb]
  \centering
  \includegraphics [width=9cm]{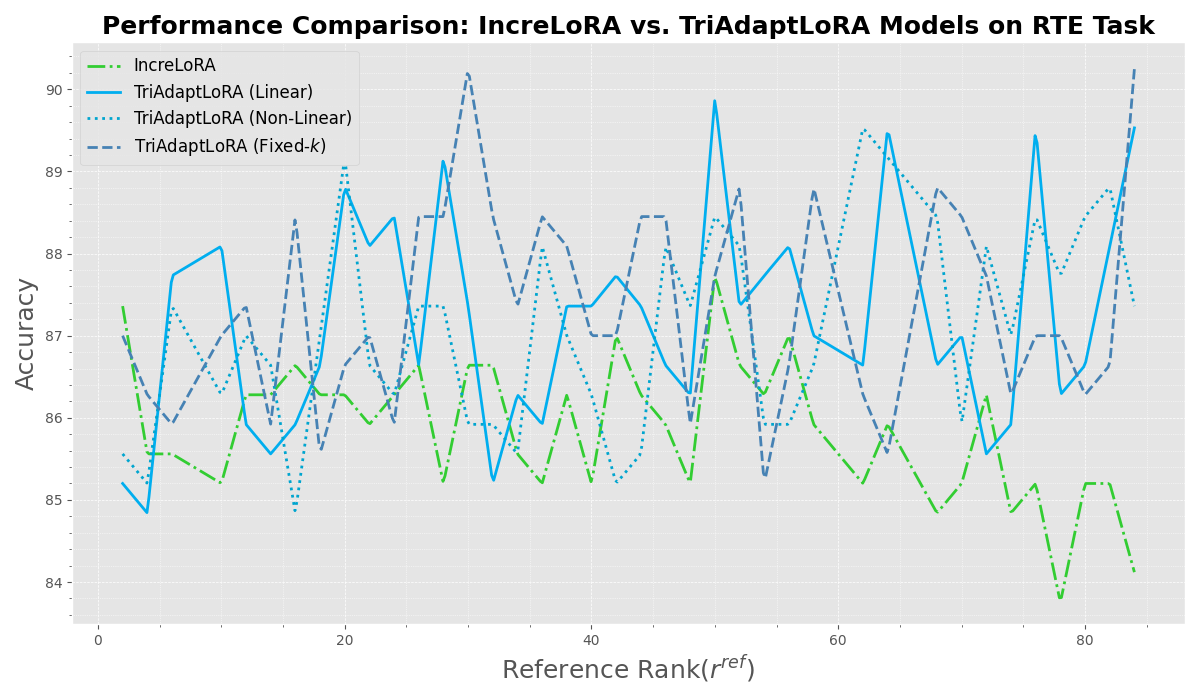}
  \caption{Experimental results on the RTE task with different reference rank sizes for TriAdaptLoRA and IncreLoRA methods based on the DeBERTaV3-base model.}
  \label{P5}
\end{figure}

The size of the reference rank ($r^{ref}$) determines the total rank budget. This experiment aims to investigate how different reference rank values ($r^{ref}$) affect model performance and stability. Accordingly, we conducted experiments on the RTE task using the DeBERTaV3-base model with two methods, TriAdaptLoRA and IncreLoRA. In these experiments, we varied the reference rank ($r^{ref}$) while keeping all other hyperparameters and random seeds consistent. Detailed experimental configurations are described in Section \ref{sec:nlu_detail_label}, and the results are illustrated in Figures \ref{P5}.

Key conclusions are as follows: 
\begin{itemize}
\item Performance: TriAdaptLoRA outperforms IncreLoRA in most reference rank ($r^{ref}$) settings, with no significant overfitting observed, further confirming its effectiveness and superiority in optimizing model performance.  
\item Impact of Different reference Ranks on Performance: As the reference rank ($r^{ref}$) increases, TriAdaptLoRA’s overall performance shows a steady improvement, suggesting that TriAdaptLoRA can leverage the additional parameters introduced by increasing the reference rank to enhance model performance. In contrast, IncreLoRA performs well at low reference ranks but its performance declines as the rank increases, indicating its inability to fully utilize the newly added parameters.
\item Stability and Parameter Efficiency: TriAdaptLoRA exhibits higher stability across different reference ranks ($r^{ref}$). Moreover, TriAdaptLoRA at lower reference ranks can already outperform IncreLoRA at certain higher ranks, indicating that simply increasing the rank (and thus the number of parameters) does not necessarily improve model performance and may even lead to performance degradation. This demonstrates that TriAdaptLoRA achieves superior efficiency in parameter utilization.
\end{itemize}

\section{Conclusion and Future Work}
This study introduces TriAdaptLoRA, a parameter-efficient fine-tuning method inspired by the brain's synaptic plasticity mechanisms. 
TriAdaptLoRA enhances model performance and reduces energy consumption by dynamically optimizing the rank allocation of incremental matrices, ensuring that matrices of higher importance receive more trainable parameters. 
Key innovations, including the triangular split of transformation matrices, importance-driven parameter allocation, and adaptive rank-growth strategies, enable TriAdaptLoRA to achieve superior parameter utilization and scalability compared to existing methods.

Comprehensive experiments demonstrate that TriAdaptLoRA consistently outperforms other PEFT methods across various natural language understanding and generation tasks, such as GLUE benchmark and SQuAD 2.0 tasks. 
Notable results include improved accuracy, reduced computational overhead, simplified hyperparameter configurations, and enhanced stability. 
These findings validate the effectiveness of TriAdaptLoRA as a robust and resource-efficient solution for fine-tuning LLMs, with significant potential for practical applications in resource-constrained environments.

While TriAdaptLoRA demonstrates promising results, further exploration is essential to fully unlock its potential. 
Future work will focus on incorporating more brain-inspired mechanisms to enhance model adaptability and parameter efficiency, leveraging principles such as long-term potentiation, spike-timing-dependent plasticity, or hierarchical processing inspired by cortical structures. 
Additionally, improving dynamic task adaptability is crucial to extend TriAdaptLoRA's application to multi-task or domain-adaptive scenarios, enabling seamless adaptation to diverse and evolving task conditions. 
Finally, advancing energy-aware optimization by explicitly integrating energy-efficiency objectives into the fine-tuning process can ensure sustainable deployment on resource-constrained hardware, aligning with broader goals of environmentally conscious AI development. Addressing these directions will further enhance the versatility, efficiency, and real-world applicability of TriAdaptLoRA.

\section*{Acknowledgments}
This work was supported in part by the Beijing Major Science and Technology Project under Contract No.Z241100001324005.

\appendices

\section{Hyperparameter Settings in Experiments}

\subsection{Natural Language Understanding Tasks}
\label{NLU_super_para}
We provide a detailed report of the hyperparameter configurations for the TriAdaptLoRA, LoRA, AdaLoRA, and IncreLoRA methods based on the DeBERTaV3-base pre-trained model, evaluated across eight datasets in the GLUE benchmark. The common hyperparameter settings and those specific to AdaLoRA are shown in Table \ref{Table5}. The first fourteen are shared hyperparameters, while the latter five are specific to the AdaLoRA method. Notably, the parameter Warm-up Steps applies only to the TriAdaptLoRA, LoRA, and IncreLoRA methods, while Incremental Rank Number and Incremental Interval are specific to the TriAdaptLoRA and IncreLoRA methods. The parameters Regularization Orthogonal Coefficient, Beta1, and Beta2 do not apply to LoRA.

\begin{table*}[htbp]
    \caption{Hyperparameter configurations for experiments using the DeBERTaV3-base model on eight GLUE benchmark tasks.}
	\label{Table5}
    \centering
    \resizebox{\textwidth}{!}{
    \begin{tabular}{l|cccccccc}
        \hline \rule{0pt}{2.5ex}\textbf{Hyper-Parameter} & \textbf{MNLI} & \textbf{SST-2} & \textbf{CoLA} & \textbf{QQP} & \textbf{QNLI} & \textbf{RTE} & \textbf{MRPC} & \textbf{STS-B} \\
        \hline
        \rule{0pt}{2.5ex}Batch Size & $32$ & $32$ & $32$ & $32$ & $32$ & $32$ & $32$ & $32$ \\
        Number of Epochs & $7$ & $24$ & $25$ & $7$ & $5$ & $50$ & $30$ & $25$ \\
        Learning Rate & $5 \mathrm{E}-04$ & $8 \mathrm{E}-04$ & $1 \mathrm{E}-03$ & $6 \mathrm{E}-04$ & $9 \mathrm{E}-04$ & $1.2 \mathrm{E}-03$ & $1 \mathrm{E}-03$ & $2.2 \mathrm{E}-03$ \\
        Incre Rank Num & $1$ & $4$ & $4$ & $4$ & $4$ & $4$ & $4$ & $4$ \\
        Weight Decay & $0$ & $0.01$ & $0$ & $0.01$ & $0.01$ & $0.01$ & $0.01$ & $0.1$ \\
        Warm-up Steps & $1000$ & $1000$ & $100$ & $1000$ & $500$ & $100$ & $100$ & $100$ \\
        Reference Rank $r^{ref}$ & $8$ & $8$ & $8$ & $8$ & $8$ & $8$ & $8$ & $8$ \\
        Alpha $\alpha$ & $16$ & $16$ & $32$ & $16$ & $32$ & $32$ & $32$ & $32$ \\
        LoRA Dropout & $0$ & $0.3$ & $default$ & $0$ & $0$ & $default$ & $0.3$ & $default$ \\
        Max Sequence Length & $256$ & $128$ & $64$ & $320$ & $512$ & $320$ & $320$ & $128$ \\
        Incre Interval & $1000$ & $1000$ & $100$ & $1000$ & $500$ & $100$ & $100$ & $100$ \\
        Reg Orth Coef & $0.1$ & $0.1$ & $0.1$ & $0.1$ & $0.1$ & $0.3$ & $0.1$ & $0.3$ \\
        Beta1 & $0.85$ & $0.85$ & $0.85$ & $0.85$ & $0.85$ & $0.85$ & $0.85$ & $0.85$ \\
        Beta2 & $0.85$ & $0.85$ & $0.85$ & $0.85$ & $0.85$ & $0.85$ & $0.85$ & $0.85$ \\
        \hline\rule{0pt}{2.5ex}Init Warm-up & $8000$ & $6000$ & $800$ & $8000$ & $2000$ & $600$ & $600$ & $800$ \\
        Final Warm-up & $50000$ & $22000$ & $3500$ & $25000$ & $8000$ & $1800$ & $1800$ & $2000$ \\
        Mask Interval & $100$ & $100$ & $10$ & $100$ & $100$ & $1$ & $1$ & $10$ \\
        Warm-up Steps/ratio & $1000$ & $1000$ & $100$ & $2000$ & $500$ & $200$ & $0.1$ & $100$ \\
        Initial LoRA rank $\mathrm{r}$ & $12$ & $12$ & $12$ & $12$ & $12$ & $12$ & $12$ & $12$ \\
        \hline
    \end{tabular}}
\end{table*}

\subsection{Natural Language Generation Tasks}
\label{NLG_super_para}

We report the hyperparameter configurations for the TriAdaptLoRA and IncreLoRA methods based on the DeBERTaV3-base pre-trained model, evaluated on the SQuAD 2.0 task. Specifically, the hyperparameter settings during the training process are detailed in Table \ref{Table6}:

\begin{table}[htbp]
	\caption{Hyperparameter configurations for experiments using the DeBERTaV3-base model on the SQuAD 2.0 task.}
	\label{Table6}
    \centering
    \small
    \setlength{\tabcolsep}{3.5pt} 
    \begin{tabular}{l|ccc}
        \hline \rule{0pt}{2.5ex}\textbf{Hyper-Parameter} & \textbf{SQuAD 2.0} \\
        \hline
        Batch Size & $16$ \\
        Number of Epochs & $14$ \\
        Learning Rate & $1.00 \mathrm{E}-03$ \\
        Incre Rank Num & $1$ \\
        Warm-up Steps & $1000$ \\
        Reference Rank $r^{ref}$ & $8$ \\
        Alpha $\alpha$ & $16$ \\
        LoRA Dropout & $0.1$ \\
        Max Sequence Length & $384$ \\
        Incre Interval & $1000$ \\
        Reg Orth Coef & $0.1$ \\
        Beta1 & $0.85$ \\
        Beta2 & $0.85$ \\
        \hline
    \end{tabular}
\end{table}

\section{Statistics of the GLUE Benchmark Tasks}
\label{glue_statistics}

The detailed statistics for the eight tasks in the GLUE benchmark are shown in Table \ref{Table7}.
\begin{table*}[htbp]
    \caption{Detailed statistical information for the eight GLUE benchmark tasks.}
	\label{Table7}
    \centering
    \resizebox{\textwidth}{!}{
    \begin{tabular}{l|cccccccc}
        \hline \rule{0pt}{2.5ex}\textbf{Hyper-Parameter} & \textbf{MNLI} & \textbf{SST-2} & \textbf{CoLA} & \textbf{QQP} & \textbf{QNLI} & \textbf{RTE} & \textbf{MRPC} & \textbf{STS-B} \\
        \hline
        \rule{0pt}{2.5ex}Batch Size & $32$ & $32$ & $32$ & $32$ & $32$ & $32$ & $32$ & $32$ \\
        Datasets & MNLI & SST-2 & CoLA & QQP & QNLI & RTE & MRPC & STS-B \\
        Train & 393k & 67k & 8.5k & 364k & 105k & 2.5k & 3.7k & 5.7k \\
        Dev & 20k & 872 & 1k & 40k & 5.5k & 276 & 408k & 1.5 k \\
        Test & 20k & 1.8k & 1k & 391k & 5.5k & 3k & 1.7k & 1.4k \\
        Metrics & Acc. & Acc. & Matthews corr. & Acc. & Acc. & Acc. & Acc. & Pearson/Spearman corr. \\
        Task & NLI & Sentiment & Acceptability & Paraphrase & QA/NLI & NLI & Paraphrase & Sentence Similarity \\
        Label & & 2 & 2 & 2 & 2 & & 2 & 1 \\
        \hline
    \end{tabular}}
\end{table*}

\bibliographystyle{IEEEtran}
\bibliography{tnnls2025}

\begin{thebibliography}{10}
\providecommand{\url}[1]{#1}
\csname url@samestyle\endcsname
\providecommand{\newblock}{\relax}
\providecommand{\bibinfo}[2]{#2}
\providecommand{\BIBentrySTDinterwordspacing}{\spaceskip=0pt\relax}
\providecommand{\BIBentryALTinterwordstretchfactor}{4}
\providecommand{\BIBentryALTinterwordspacing}{\spaceskip=\fontdimen2\font plus
\BIBentryALTinterwordstretchfactor\fontdimen3\font minus \fontdimen4\font\relax}
\providecommand{\BIBforeignlanguage}[2]{{%
\expandafter\ifx\csname l@#1\endcsname\relax
\typeout{** WARNING: IEEEtran.bst: No hyphenation pattern has been}%
\typeout{** loaded for the language `#1'. Using the pattern for}%
\typeout{** the default language instead.}%
\else
\language=\csname l@#1\endcsname
\fi
#2}}
\providecommand{\BIBdecl}{\relax}
\BIBdecl

\bibitem{ouyang2022training}
L.~Ouyang, J.~Wu, X.~Jiang, D.~Almeida, C.~Wainwright, P.~Mishkin, C.~Zhang, S.~Agarwal, K.~Slama, A.~Ray \emph{et~al.}, ``Training language models to follow instructions with human feedback,'' \emph{Advances in Neural Information Processing Systems}, vol.~35, pp. 27\,730--27\,744, 2022.

\bibitem{ding2023parameter}
N.~Ding, Y.~Qin, G.~Yang, F.~Wei, Z.~Yang, Y.~Su, S.~Hu, Y.~Chen, C.-M. Chan, W.~Chen \emph{et~al.}, ``Parameter-efficient fine-tuning of large-scale pre-trained language models,'' \emph{Nature Machine Intelligence}, vol.~5, no.~3, pp. 220--235, 2023.

\bibitem{ziems-etal-2023-large}
\BIBentryALTinterwordspacing
N.~Ziems, W.~Yu, Z.~Zhang, and M.~Jiang, ``Large language models are built-in autoregressive search engines,'' in \emph{Findings of the Association for Computational Linguistics: ACL 2023}, A.~Rogers, J.~Boyd-Graber, and N.~Okazaki, Eds.\hskip 1em plus 0.5em minus 0.4em\relax Toronto, Canada: Association for Computational Linguistics, Jul. 2023, pp. 2666--2678. [Online]. Available: \url{https://aclanthology.org/2023.findings-acl.167}
\BIBentrySTDinterwordspacing

\bibitem{achiam2023gpt}
J.~Achiam, S.~Adler, S.~Agarwal, L.~Ahmad, I.~Akkaya, F.~L. Aleman, D.~Almeida, J.~Altenschmidt, S.~Altman, S.~Anadkat \emph{et~al.}, ``Gpt-4 technical report,'' \emph{arXiv preprint arXiv:2303.08774}, 2023.

\bibitem{dubey2024llama}
A.~Dubey, A.~Jauhri, A.~Pandey, A.~Kadian, A.~Al-Dahle, A.~Letman, A.~Mathur, A.~Schelten, A.~Yang, A.~Fan \emph{et~al.}, ``The llama 3 herd of models,'' \emph{arXiv preprint arXiv:2407.21783}, 2024.

\bibitem{touvron2023llama2}
H.~Touvron, L.~Martin, K.~Stone, P.~Albert, A.~Almahairi, Y.~Babaei, N.~Bashlykov, S.~Batra, P.~Bhargava, S.~Bhosale \emph{et~al.}, ``Llama 2: Open foundation and fine-tuned chat models,'' \emph{arXiv preprint arXiv:2307.09288}, 2023.

\bibitem{askell2021general}
A.~Askell, Y.~Bai, A.~Chen, D.~Drain, D.~Ganguli, T.~Henighan, A.~Jones, N.~Joseph, B.~Mann, N.~DasSarma \emph{et~al.}, ``A general language assistant as a laboratory for alignment,'' \emph{arXiv preprint arXiv:2112.00861}, 2021.

\bibitem{wei2021finetuned}
J.~Wei, M.~Bosma, V.~Y. Zhao, K.~Guu, A.~W. Yu, B.~Lester, N.~Du, A.~M. Dai, and Q.~V. Le, ``Finetuned language models are zero-shot learners,'' \emph{arXiv preprint arXiv:2109.01652}, 2021.

\bibitem{min2021metaicl}
S.~Min, M.~Lewis, L.~Zettlemoyer, and H.~Hajishirzi, ``Metaicl: Learning to learn in context,'' \emph{arXiv preprint arXiv:2110.15943}, 2021.

\bibitem{liu2022few}
H.~Liu, D.~Tam, M.~Muqeeth, J.~Mohta, T.~Huang, M.~Bansal, and C.~A. Raffel, ``Few-shot parameter-efficient fine-tuning is better and cheaper than in-context learning,'' \emph{Advances in Neural Information Processing Systems}, vol.~35, pp. 1950--1965, 2022.

\bibitem{qiu2020pre}
X.~Qiu, T.~Sun, Y.~Xu, Y.~Shao, N.~Dai, and X.~Huang, ``Pre-trained models for natural language processing: A survey,'' \emph{Science China Technological Sciences}, vol.~63, no.~10, pp. 1872--1897, 2020.

\bibitem{raffel2020exploring}
C.~Raffel, N.~Shazeer, A.~Roberts, K.~Lee, S.~Narang, M.~Matena, Y.~Zhou, W.~Li, and P.~J. Liu, ``Exploring the limits of transfer learning with a unified text-to-text transformer,'' \emph{The Journal of Machine Learning Research}, vol.~21, no.~1, pp. 5485--5551, 2020.

\bibitem{hu2022lora}
\BIBentryALTinterwordspacing
E.~J. Hu, Y.~Shen, P.~Wallis, Z.~Allen-Zhu, Y.~Li, S.~Wang, L.~Wang, and W.~Chen, ``Lo{RA}: Low-rank adaptation of large language models,'' in \emph{International Conference on Learning Representations}, 2022. [Online]. Available: \url{https://openreview.net/forum?id=nZeVKeeFYf9}
\BIBentrySTDinterwordspacing

\bibitem{liu2024dora}
S.-Y. Liu, C.-Y. Wang, H.~Yin, P.~Molchanov, Y.-C.~F. Wang, K.-T. Cheng, and M.-H. Chen, ``Dora: Weight-decomposed low-rank adaptation,'' \emph{arXiv preprint arXiv:2402.09353}, 2024.

\bibitem{kopiczko_vera_2024}
\BIBentryALTinterwordspacing
D.~J. Kopiczko, T.~Blankevoort, and Y.~M. Asano, ``{VeRA}: {Vector}-based {Random} {Matrix} {Adaptation},'' in \emph{The {Twelfth} {International} {Conference} on {Learning} {Representations}}, 2024. [Online]. Available: \url{https://openreview.net/forum?id=NjNfLdxr3A}
\BIBentrySTDinterwordspacing

\bibitem{liu_aflora_2024}
\BIBentryALTinterwordspacing
Z.~Liu, S.~Kundu, A.~Li, J.~Wan, L.~Jiang, and P.~Beerel, ``\BIBforeignlanguage{en}{{AFLoRA}: {Adaptive} {Freezing} of {Low} {Rank} {Adaptation} in {Parameter} {Efficient} {Fine}-{Tuning} of {Large} {Models}},'' in \emph{\BIBforeignlanguage{en}{Proceedings of the 62nd {Annual} {Meeting} of the {Association} for {Computational} {Linguistics} ({Volume} 2: {Short} {Papers})}}, L.-W. Ku, A.~Martins, and V.~Srikumar, Eds.\hskip 1em plus 0.5em minus 0.4em\relax Bangkok, Thailand: Association for Computational Linguistics, Aug. 2024, pp. 161--167. [Online]. Available: \url{https://aclanthology.org/2024.acl-short.16}
\BIBentrySTDinterwordspacing

\bibitem{wang_prolora_2024}
\BIBentryALTinterwordspacing
S.~Wang, B.~Xue, J.~Ye, J.~Jiang, L.~Chen, L.~Kong, and C.~Wu, ``\BIBforeignlanguage{en}{{PRoLoRA}: {Partial} {Rotation} {Empowers} {More} {Parameter}-{Efficient} {LoRA}},'' in \emph{\BIBforeignlanguage{en}{Proceedings of the 62nd {Annual} {Meeting} of the {Association} for {Computational} {Linguistics} ({Volume} 1: {Long} {Papers})}}, L.-W. Ku, A.~Martins, and V.~Srikumar, Eds.\hskip 1em plus 0.5em minus 0.4em\relax Bangkok, Thailand: Association for Computational Linguistics, Aug. 2024, pp. 2829--2841. [Online]. Available: \url{https://aclanthology.org/2024.acl-long.156}
\BIBentrySTDinterwordspacing

\bibitem{chen2023longlora}
Y.~Chen, S.~Qian, H.~Tang, X.~Lai, Z.~Liu, S.~Han, and J.~Jia, ``Longlora: Efficient fine-tuning of long-context large language models,'' \emph{arXiv preprint arXiv:2309.12307}, 2023.

\bibitem{dettmers2024qlora}
T.~Dettmers, A.~Pagnoni, A.~Holtzman, and L.~Zettlemoyer, ``Qlora: Efficient finetuning of quantized llms,'' \emph{Advances in Neural Information Processing Systems}, vol.~36, 2024.

\bibitem{wang_lora-flow_2024}
\BIBentryALTinterwordspacing
H.~Wang, B.~Ping, S.~Wang, X.~Han, Y.~Chen, Z.~Liu, and M.~Sun, ``\BIBforeignlanguage{en}{{LoRA}-{Flow}: {Dynamic} {LoRA} {Fusion} for {Large} {Language} {Models} in {Generative} {Tasks}},'' in \emph{\BIBforeignlanguage{en}{Proceedings of the 62nd {Annual} {Meeting} of the {Association} for {Computational} {Linguistics} ({Volume} 1: {Long} {Papers})}}, L.-W. Ku, A.~Martins, and V.~Srikumar, Eds.\hskip 1em plus 0.5em minus 0.4em\relax Bangkok, Thailand: Association for Computational Linguistics, Aug. 2024, pp. 12\,871--12\,882. [Online]. Available: \url{https://aclanthology.org/2024.acl-long.695}
\BIBentrySTDinterwordspacing

\bibitem{zhang2023adaptive}
Q.~Zhang, M.~Chen, A.~Bukharin, P.~He, Y.~Cheng, W.~Chen, and T.~Zhao, ``Adaptive budget allocation for parameter-efficient fine-tuning,'' \emph{arXiv preprint arXiv:2303.10512}, 2023.

\bibitem{Zhang2023IncreLoRAIP}
\BIBentryALTinterwordspacing
F.~F. Zhang, L.~Li, J.-C. Chen, Z.~Jiang, B.~Wang, and Y.~Qian, ``Increlora: Incremental parameter allocation method for parameter-efficient fine-tuning,'' \emph{ArXiv}, vol. abs/2308.12043, 2023. [Online]. Available: \url{https://api.semanticscholar.org/CorpusID:261076438}
\BIBentrySTDinterwordspacing

\bibitem{song2024increasing}
\BIBentryALTinterwordspacing
H.~SONG, H.~Zhao, S.~Majumder, and T.~Lin, ``Increasing model capacity for free: A simple strategy for parameter efficient fine-tuning,'' in \emph{The Twelfth International Conference on Learning Representations}, 2024. [Online]. Available: \url{https://openreview.net/forum?id=H3IUunLy8s}
\BIBentrySTDinterwordspacing

\bibitem{hebb_organization_2005}
D.~O. Hebb, \emph{The organization of behavior: {A} neuropsychological theory}.\hskip 1em plus 0.5em minus 0.4em\relax Psychology press, 2005.

\bibitem{wang2018glue}
A.~Wang, A.~Singh, J.~Michael, F.~Hill, O.~Levy, and S.~R. Bowman, ``Glue: A multi-task benchmark and analysis platform for natural language understanding,'' \emph{arXiv preprint arXiv:1804.07461}, 2018.

\bibitem{rajpurkar-etal-2018-know}
\BIBentryALTinterwordspacing
P.~Rajpurkar, R.~Jia, and P.~Liang, ``Know what you don{'}t know: Unanswerable questions for {SQ}u{AD},'' in \emph{Proceedings of the 56th Annual Meeting of the Association for Computational Linguistics (Volume 2: Short Papers)}, I.~Gurevych and Y.~Miyao, Eds.\hskip 1em plus 0.5em minus 0.4em\relax Melbourne, Australia: Association for Computational Linguistics, Jul. 2018, pp. 784--789. [Online]. Available: \url{https://aclanthology.org/P18-2124}
\BIBentrySTDinterwordspacing

\bibitem{vaswani2017attention}
A.~Vaswani, N.~Shazeer, N.~Parmar, J.~Uszkoreit, L.~Jones, A.~N. Gomez, {\L}.~Kaiser, and I.~Polosukhin, ``Attention is all you need,'' \emph{Advances in neural information processing systems}, vol.~30, 2017.

\bibitem{valipour_dylora_2023}
\BIBentryALTinterwordspacing
M.~Valipour, M.~Rezagholizadeh, I.~Kobyzev, and A.~Ghodsi, ``\BIBforeignlanguage{en}{{DyLoRA}: {Parameter}-{Efficient} {Tuning} of {Pre}-trained {Models} using {Dynamic} {Search}-{Free} {Low}-{Rank} {Adaptation}},'' in \emph{\BIBforeignlanguage{en}{Proceedings of the 17th {Conference} of the {European} {Chapter} of the {Association} for {Computational} {Linguistics}}}.\hskip 1em plus 0.5em minus 0.4em\relax Dubrovnik, Croatia: Association for Computational Linguistics, 2023, pp. 3274--3287. [Online]. Available: \url{https://aclanthology.org/2023.eacl-main.239}
\BIBentrySTDinterwordspacing

\bibitem{littman_reinforcement_2015}
M.~L. Littman, ``\BIBforeignlanguage{en}{Reinforcement learning improves behaviour from evaluative feedback},'' \emph{\BIBforeignlanguage{en}{Nature}}, vol. 521, no. 7553, pp. 445--451, 2015, publisher: Nature Publishing Group UK London.

\bibitem{roelfsema_control_2018}
\BIBentryALTinterwordspacing
P.~R. Roelfsema and A.~Holtmaat, ``\BIBforeignlanguage{en}{Control of synaptic plasticity in deep cortical networks},'' \emph{\BIBforeignlanguage{en}{Nature Reviews Neuroscience}}, vol.~19, no.~3, pp. 166--180, Mar. 2018. [Online]. Available: \url{https://www.nature.com/articles/nrn.2018.6}
\BIBentrySTDinterwordspacing

\bibitem{sutton_reinforcement_2018}
R.~S. Sutton, ``\BIBforeignlanguage{en}{Reinforcement learning: {An} introduction},'' \emph{\BIBforeignlanguage{en}{A Bradford Book}}, 2018.

\bibitem{he2023debertav}
\BIBentryALTinterwordspacing
P.~He, J.~Gao, and W.~Chen, ``De{BERT}av3: Improving de{BERT}a using {ELECTRA}-style pre-training with gradient-disentangled embedding sharing,'' in \emph{The Eleventh International Conference on Learning Representations}, 2023. [Online]. Available: \url{https://openreview.net/forum?id=sE7-XhLxHA}
\BIBentrySTDinterwordspacing

\bibitem{rebuffi2017learning}
S.-A. Rebuffi, H.~Bilen, and A.~Vedaldi, ``Learning multiple visual domains with residual adapters,'' \emph{Advances in neural information processing systems}, vol.~30, 2017.

\bibitem{houlsby2019parameter}
N.~Houlsby, A.~Giurgiu, S.~Jastrzebski, B.~Morrone, Q.~De~Laroussilhe, A.~Gesmundo, M.~Attariyan, and S.~Gelly, ``Parameter-efficient transfer learning for nlp,'' in \emph{International Conference on Machine Learning}.\hskip 1em plus 0.5em minus 0.4em\relax PMLR, 2019, pp. 2790--2799.

\bibitem{he2022towards}
\BIBentryALTinterwordspacing
J.~He, C.~Zhou, X.~Ma, T.~Berg-Kirkpatrick, and G.~Neubig, ``Towards a unified view of parameter-efficient transfer learning,'' in \emph{International Conference on Learning Representations}, 2022. [Online]. Available: \url{https://openreview.net/forum?id=0RDcd5Axok}
\BIBentrySTDinterwordspacing

\bibitem{zaken2021bitfit}
E.~B. Zaken, S.~Ravfogel, and Y.~Goldberg, ``Bitfit: Simple parameter-efficient fine-tuning for transformer-based masked language-models,'' \emph{arXiv preprint arXiv:2106.10199}, 2021.

\bibitem{warstadt2019neural}
A.~Warstadt, A.~Singh, and S.~R. Bowman, ``Neural network acceptability judgments,'' \emph{Transactions of the Association for Computational Linguistics}, vol.~7, pp. 625--641, 2019.

\bibitem{matthews1975comparison}
B.~W. Matthews, ``Comparison of the predicted and observed secondary structure of t4 phage lysozyme,'' \emph{Biochimica et Biophysica Acta (BBA)-Protein Structure}, vol. 405, no.~2, pp. 442--451, 1975.

\bibitem{williams2017broad}
A.~Williams, N.~Nangia, and S.~R. Bowman, ``A broad-coverage challenge corpus for sentence understanding through inference,'' \emph{arXiv preprint arXiv:1704.05426}, 2017.

\bibitem{dolan2005automatically}
B.~Dolan and C.~Brockett, ``Automatically constructing a corpus of sentential paraphrases,'' in \emph{Third International Workshop on Paraphrasing (IWP2005)}, 2005.

\bibitem{rajpurkar2016squad}
P.~Rajpurkar, J.~Zhang, K.~Lopyrev, and P.~Liang, ``Squad: 100,000+ questions for machine comprehension of text,'' \emph{arXiv preprint arXiv:1606.05250}, 2016.

\bibitem{dagan2005pascal}
I.~Dagan, O.~Glickman, and B.~Magnini, ``The pascal recognising textual entailment challenge,'' in \emph{Machine learning challenges workshop}.\hskip 1em plus 0.5em minus 0.4em\relax Springer, 2005, pp. 177--190.

\bibitem{haim2006second}
R.~B. Haim, I.~Dagan, B.~Dolan, L.~Ferro, D.~Giampiccolo, B.~Magnini, and I.~Szpektor, ``The second pascal recognising textual entailment challenge,'' in \emph{Proceedings of the Second PASCAL Challenges Workshop on Recognising Textual Entailment}, vol.~7, 2006, pp. 785--794.

\bibitem{giampiccolo2007third}
D.~Giampiccolo, B.~Magnini, I.~Dagan, and W.~B. Dolan, ``The third pascal recognizing textual entailment challenge,'' in \emph{Proceedings of the ACL-PASCAL workshop on textual entailment and paraphrasing}, 2007, pp. 1--9.

\bibitem{bentivogli2009fifth}
L.~Bentivogli, P.~Clark, I.~Dagan, and D.~Giampiccolo, ``The fifth pascal recognizing textual entailment challenge.'' \emph{TAC}, vol.~7, p.~8, 2009.

\bibitem{socher2013recursive}
R.~Socher, A.~Perelygin, J.~Wu, J.~Chuang, C.~D. Manning, A.~Y. Ng, and C.~Potts, ``Recursive deep models for semantic compositionality over a sentiment treebank,'' in \emph{Proceedings of the 2013 conference on empirical methods in natural language processing}, 2013, pp. 1631--1642.

\bibitem{cer2017semeval}
D.~Cer, M.~Diab, E.~Agirre, I.~Lopez-Gazpio, and L.~Specia, ``Semeval-2017 task 1: Semantic textual similarity-multilingual and cross-lingual focused evaluation,'' \emph{arXiv preprint arXiv:1708.00055}, 2017.

\end{thebibliography}

\end{document}